\documentclass{article}

\usepackage{multirow}
\usepackage{microtype}
\usepackage{graphicx}
\usepackage{subfigure}
\usepackage{booktabs} 
\usepackage{hyperref}

\usepackage[accepted]{icml2025}

\usepackage{amsmath,amssymb,mathtools,amsthm}
\usepackage[capitalize,noabbrev]{cleveref}

\usepackage[textsize=tiny]{todonotes}

\icmltitlerunning{Predicting Cellular Composition from H\&E Images}

\begin{document}

\twocolumn[

\icmltitle{Integrating Pathology Foundation Models and Spatial Transcriptomics for Cellular Decomposition from Histology Images}

\begin{icmlauthorlist}
\icmlauthor{Yutong Sun}{ece}
\icmlauthor{Sichen Zhu}{bme}
\icmlauthor{Peng Qiu}{bme}
\end{icmlauthorlist}

\icmlaffiliation{ece}{School of Electrical and Computer Engineering, Georgia Institute of Technology, Atlanta, GA, USA}
\icmlaffiliation{bme}{Department of Biomedical Engineering, Georgia Institute of Technology and Emory University, Atlanta, GA, USA}

\icmlcorrespondingauthor{Peng Qiu}{peng.qiu@bme.gatech.edu}

\vskip 0.3in
]

\printAffiliationsAndNotice{}

\begin{abstract}
The rapid development of digital pathology and modern deep learning has facilitated the emergence of pathology foundation models that are expected to solve general pathology problems under various disease conditions in one unified model, with or without fine-tuning. In parallel, spatial transcriptomics has emerged as a transformative technology that enables the profiling of gene expression on hematoxylin and eosin (H\&E) stained histology images. Spatial transcriptomics unlocks the unprecedented opportunity to dive into existing histology images at a more granular, cellular level. In this work, we propose a lightweight and training-efficient approach to predict cellular composition directly from H\&E-stained histology images by leveraging information-enriched feature embeddings extracted from pre-trained pathology foundation models. By training a lightweight multi-layer perceptron (MLP) regressor on cell-type abundances derived via cell2location, our method efficiently distills knowledge from pathology foundation models and demonstrates the ability to accurately predict cell-type compositions from histology images, without physically performing the costly spatial transcriptomics. Our method demonstrates competitive performance compared to existing methods such as Hist2Cell, while significantly reducing computational complexity.
\end{abstract}

\section{Introduction}
\label{sec:aim3_intro}

Spatial transcriptomics (ST) enables comprehensive gene expression analysis while preserving its spatial context. In 2021, \textit{Nature Methods} named it the “Method of the Year 2020,” underscoring its impact. Among available platforms, 10x Genomics Visium is widely used for whole-genome expression profiling. Visium uses 55$\mu \mathrm{m}$ spots on hematoxylin and eosin (H\&E) stained tissue slides, each capturing signals from multiple cells within the spot and achieving a sub-optimal to single cell resolution. 

Given the gene expression for each spot, spatial deconvolution algorithms infer cell-type composition per spot by integrating spatially resolved gene expressions and scRNA-seq references from other independent studies. Existing methods like cell2location \cite{kleshchevnikov2022cell2location}, RCTD \cite{cable2022robust}, and Stereoscope \cite{andersson2020single} use probabilistic models to model the distribution of spatially resolved gene expression, while matrix factorization techniques such as SpatialDWLS \cite{dong2021spatialdwls} and SPOTlight \cite{elosua2021spotlight} estimate proportions via regression. Other methods like CARD \cite{ma2022spatially} and deep learning models such as Tangram \cite{biancalani2021deep} and DestVI \cite{lopez2022destvi} incorporate spatial priors or align with single-cell references.

Those approaches mentioned above require ST data and high-quality single-cell references to obtain cellular compositions on H\&E stained histology images. However, histology images themselves carry substantial information, such as cell morphology and tissue structure, for cell type identification. To reduce the burden of performing ST to obtain quantitative cellular-level information, image-based methods have emerged to predict cellular compositions at the $\mu \mathrm{m}$-scale resolution. Hist2Cell, for example, uses Vision Graph-Transformers to infer cell types directly from H\&E images but remains computationally intensive.

Recently, pathology foundation models, such as CONCH \cite{lu2024visual} and UNI \cite{chen2024towards}, have demonstrated great versatility by extracting high-quality morphological embeddings from large cohorts of H\&E training images and have achieved success in some downstream tasks \cite{zhu2025diffusion}. When paired with lightweight multi-layer perceptron (MLP) regressors, they provide the opportunity for efficient cell type proportion predictions without complicated neural network architectures and time-consuming training protocols. 

Therefore, we propose a cell type prediction pipeline that eliminates the need for gene expression data. We train an MLP regressor on cell2location-derived ground truths using pre-trained embeddings from pathology foundation models. Our method achieves performance on par with, or better than existing image-based methods, while offering significant improvements in computational efficiency and scalability.

\section{Methods and Approach}
\label{sec:aim3_methods}

\subsection{Overview}
\label{subsec:workflow_overview}

\begin{figure*}[ht]
    \centering
    { 
    \resizebox{0.7\textwidth}{!}{\includegraphics{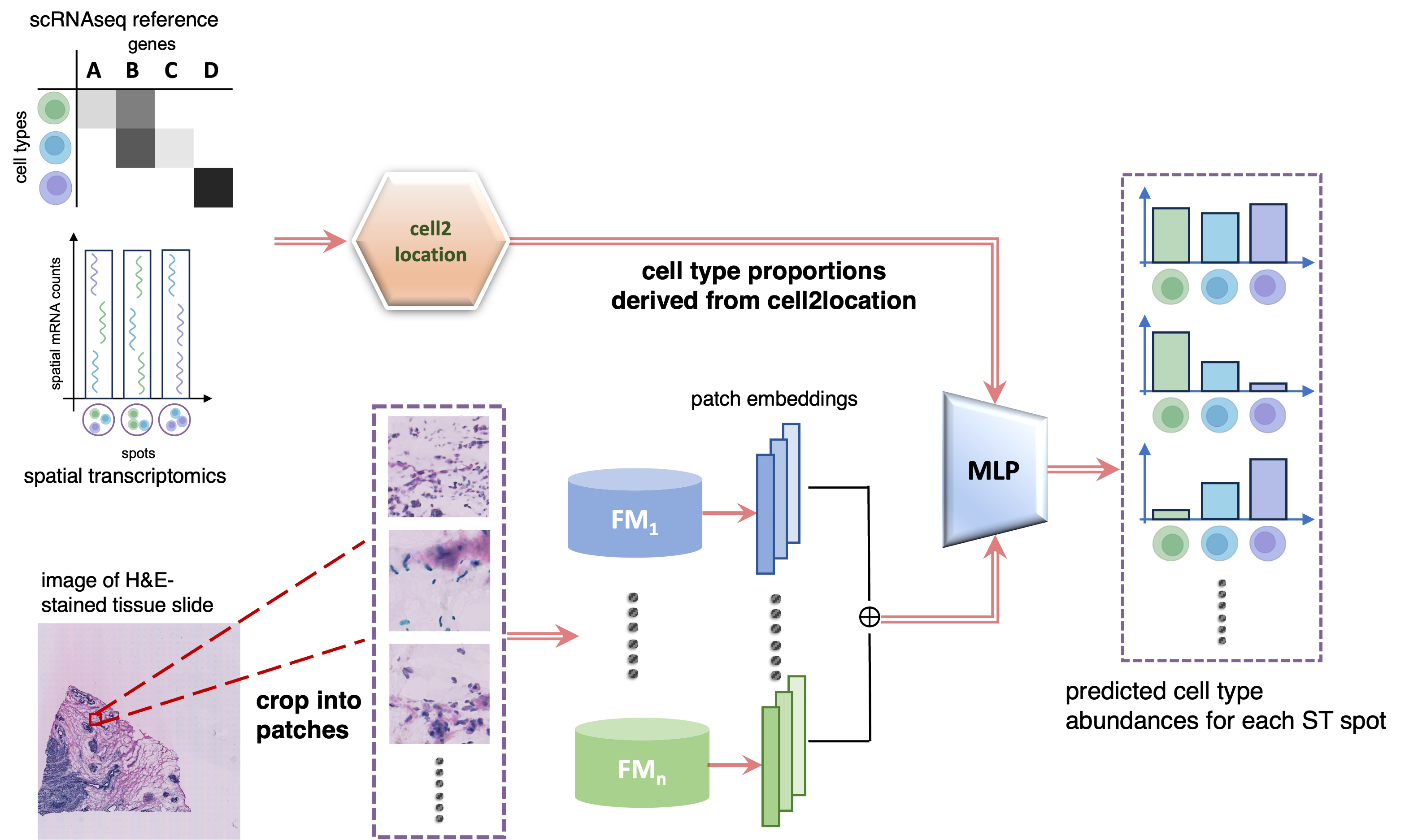}}
    \caption[Overview of the cell type abundance prediction pipeline]{\textbf{Overview of the prediction pipeline.} The H\&E slide is cropped into patches at spatial spots and passed through multiple pre-trained models to extract and concatenate embeddings, which are input to an MLP regressor for cell type prediction. In parallel, cell2location integrates gene expression and scRNA-seq data to generate ground truth labels for training and evaluation.}
    \label{fig:spatial_workflow}}
\end{figure*}

The proposed pipeline aims to infer cell-type abundances for each spot on H\&E-stained histology images by leveraging high-quality embeddings extracted from pre-trained foundation models. As illustrated in \cref{fig:spatial_workflow}, ST data consists of both a spot-level gene count matrix and an H\&E-stained tissue slide. In our approach, we exclusively utilize the H\&E-stained tissue slide, which is cropped into image patches centered on the ST spots.

Each image patch is passed through multiple pre-trained foundation models, denoted as \( FM_{1}, ..., FM_{n} \), generating \( n \) embeddings, which are concatenated into a single feature vector. In parallel, we run cell2location on the spatial gene count matrix and scRNA-seq reference to estimate spot-level ground truth abundances for model training and evaluation. Our MLP regressor takes the feature vector for the image patch as input and then outputs the predicted cell type abundance for the spot. 

To supervise training, we define a composite loss function:
{
\( L_{\text{total}} = \text{MSE}(\hat{y}, y) + \lambda_1 \cdot \text{MAE}(\hat{y}, y) + \lambda_2 \cdot L_{\text{Pearson}}(\hat{y}, y) \)},
where \( \hat{y} \) and \( y \) denote predicted and ground truth cell-type abundances, and \( \lambda_1, \lambda_2 \) are weighting coefficients. The MSE term penalizes large deviations, MAE enhances robustness to outliers, and \textbf{\( L_{\text{Pearson}} \)} promotes strong linear correlation by aligning relative abundance ranks:

{
\( L_{\text{Pearson}} = -\frac{ \sum (\hat{y}_i - \bar{\hat{y}})(y_i - \bar{y}) }{ \sqrt{\sum (\hat{y}_i - \bar{\hat{y}})^2} \cdot \sqrt{\sum (y_i - \bar{y})^2} + \epsilon } \)},
where \( \bar{\hat{y}} \) and \( \bar{y} \) are the means of predictions and ground truth, and \( \epsilon \) ensures numerical stability. 

This approach cuts computational cost, allows quick inference at test time, and generalizes across tissue types by leveraging foundation model embeddings that capture diverse morphology from a large training cohort.

\noindent
\subsection{Embedding Concatenation and MLP Regression}
\label{subsec:embedding_MLP}

We evaluate three embedding strategies: (1) CONCH, (2) UNI, and (3) concatenated CONCH+UNI. CONCH is a vision-language model, while UNI is a purely vision model. The combined embedding in (3) captures complementary features. The embeddings are input to a 7-layer MLP with SiLU activations and a final linear layer to predict cell-type abundances. All embeddings are standardized (zero mean, unit variance) prior to training.

\subsection{Data Preprocessing}

We use two breast cancer ST datasets: 1) her2st \cite{andersson2021spatial}, including HER2-positive tumors from eight patients (36 sections, 13,620 spots); 2) STNet \cite{he2020integrating}, which covers 23 patients (30,655 spots). Cell-type abundances derived via cell2location \cite{kleshchevnikov2022cell2location} using the Human Breast Cell Atlas (HBCA) \cite{kumar2023spatially} as reference ensure consistent ground truth labels.

Image preprocessing follows the Hist2Cell pipeline \cite{zhao2024hist2cell}. We cropped $224 \times 224$ patches centered on each ST spot. Background-dominated patches identified by RGB thresholds are excluded to focus training on informative tissue regions.

\subsection{Experimental Outline}
We evaluate the model using two setups to assess robustness and generalizability. The first is leave-one-out cross-validation, where the model is trained on all but one patient and tested on the held-out sample. This process is repeated for each patient in the dataset. In the second setting, we evaluate cross-dataset generalization by training the model on one dataset (e.g., her2st) and testing on another (e.g., STNet). Under such out-of-distribution setting, we hope to measure model's transferability across diverse histological and molecular profiles.

We assess performance using Pearson correlation (CC Score) and L1 Score to evaluate the accuracy of cell-type abundance predictions. Additionally, to capture spatial organization in the tissue, we analyze colocalization patterns using bivariate Moran’s R, along with cosine similarity and correlation between the predicted and ground truth colocalization maps.

To evaluate computational efficiency, we compare GPU memory usage and training time across different algorithms. By leveraging pretrained image embeddings, our method significantly reduces computational cost compared to training deep feature-extracting models from scratch, without compromising accuracy.

\section{Results}
\label{sec:aim3_results}

\subsection{Cell Type Abundance Prediction}
\label{sec:cell_type_abundance}

\begin{figure}[ht]
    \centering    
    { 
    \includegraphics[width=\columnwidth]{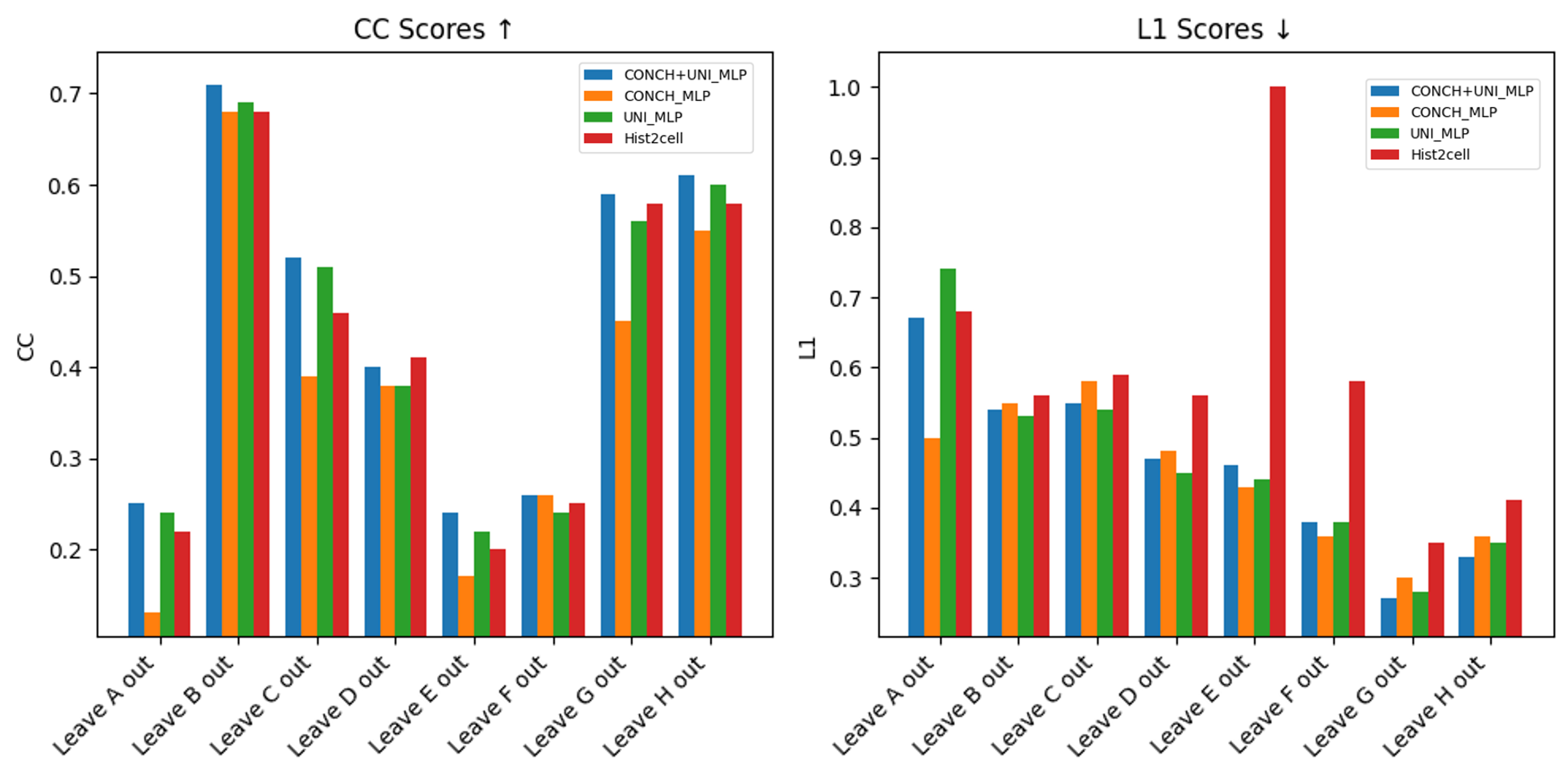}
    \caption[Cell type abundance prediction on her2st]{\textbf{Cell type abundance prediction on the her2st dataset.} (\textbf{Left}) CC scores; (\textbf{Right}) L1 errors. The CONCH+UNI\_MLP model outperforms individual embeddings and Hist2Cell in most leave-one-out cases.}
    \label{fig:cc_l1_her2st}}
\end{figure}

\begin{figure}[ht]
    \centering
    {     \includegraphics[width=1\columnwidth]{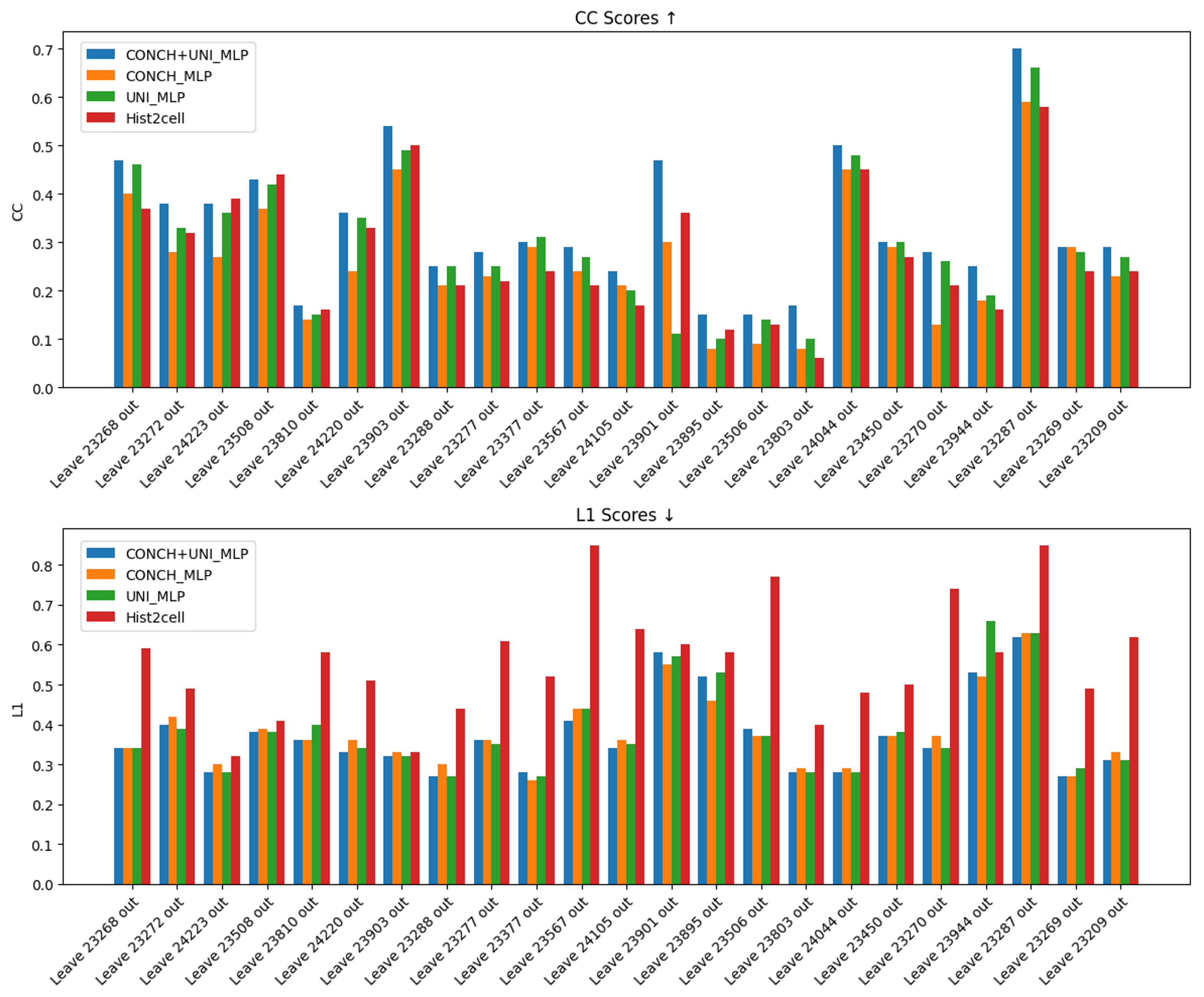}
    \caption[Cell type abundance prediction on STNet]{\textbf{Cell type abundance prediction on the STNet dataset.} (\textbf{Top}) CC scores; (\textbf{Bottom}) L1 errors. The CONCH+UNI\_MLP model consistently shows better performance.}
    \label{fig:cc_l1_stnet}}
\end{figure}

To the best of our current knowledge, Hist2Cell is the only existing method in literature that predicts cell-type abundance from histology images. So, Hist2Cell is the main method that we compare our pipeline with. \cref{fig:cc_l1_her2st,fig:cc_l1_stnet} compare CC and L1 scores for our approach versus Hist2Cell. Higher CC and lower L1 scores indicate better performance, prioritizing CC scores for their biological relevance. Our method generally outperforms Hist2Cell, with exceptions in one her2st sample (D) and two STNet samples (24223, 23508). Concatenated foundation model embeddings outperform single-model embeddings in all but two STNet samples (23288, 23377). We hypothesized that combined features from CONCH and UNI would benefit from complementary information captured by different pretraining techniques and different modalities.  

We notice that the results from UNI embeddings alone surpass CONCH embeddings' results, likely due to UNI’s extensive histopathology-focused training. However, concatenating CONCH and UNI further improves performance, which could be attributed to integrating visual and textual features.

Visualization of cell-type predictions on her2st slide B1 (\cref{fig:B1}(a)) highlights four representative cell types: luminal epithelial cells, plasmacytoid dendritic cells, proliferating myeloid cells, and capillary endothelial cells. Luminal epithelial cells exhibit high abundance consistently, while plasmacytoid dendritic cells vary notably between training (10th) and testing (25th). Proliferating myeloid cells and capillary endothelial cells are consistently among the least abundant. Our pipeline effectively captures spatial distributions and significantly reduces false positives compared to Hist2Cell, achieving an average CC improvement of 0.1, particularly benefiting cell types with less distinct morphological patterns.

\subsection{Cell Type Colocalization}
\label{sec:colocalization}

Moran’s \( R \) measures spatial colocalization between two cell types and is defined as
{
\( R = \frac{\sum_i \sum_j w_{ij} (x_i - \bar{x}) (y_j - \bar{y})}
{\sqrt{\sum_i (x_i - \bar{x})^2 \sum_i (y_i - \bar{y})^2}} \),
}
where \( x_i \) and \( y_j \) denote cell-type abundances at locations \( i \) and \( j \), respectively, and \( w_{ij} \) is a spatial weight computed using a radial basis function (RBF) kernel:
{
\( w_{ij}^{(0)} = \exp \left( -\frac{d_{ij}^2}{2l^2} \right), \quad
w_{ij} = \frac{n}{W} w_{ij}^{(0)} \),
}
with \( d_{ij} \) the Euclidean distance between spots, {\( W = \sum w_{ij}^{(0)} \)}, and \( n \) the total number of spots.

\cref{fig:B1}(b) shows colocalization clustermaps from ground truth, our method, and Hist2Cell predictions, demonstrating our method's comparable spatial consistency.

Quantitative evaluation using row-wise cosine similarity and correlation between the colocalization maps predicted by each method and the ground truth (\cref{tab:colocalization_her2st_stnet}) confirms our method matches Hist2Cell performance across both datasets. Despite Hist2Cell explicitly modeling spatial context via Graph-Transformer architectures, our approach achieves similar results through concatenated embeddings from CONCH and UNI, indicating these embeddings capture spatially relevant morphological features and are capable of recovering corresponding cell-type abundance.

\begin{table}[t]
    \centering
    { 
    \caption[Quantitative comparison of colocalization performance]{\textbf{Quantitative comparison of cell-type colocalization performance.} Row-wise cosine similarity and correlation between predicted and ground truth colocalization maps for the her2st and STNet datasets.}
    \label{tab:colocalization_her2st_stnet}

    \resizebox{\columnwidth}{!}{%
    \renewcommand{\arraystretch}{1.3}
    \begin{tabular}{|l|c|c|c|c|}
        \hline
        \multirow{2}{*}{\textbf{Model}} & \multicolumn{2}{c|}{\textbf{Her2st}} & \multicolumn{2}{c|}{\textbf{STNet}} \\
        \cline{2-5}
         & \textbf{Cosine Similarity} & \textbf{Correlation} & \textbf{Cosine Similarity} & \textbf{Correlation} \\
        \hline
        CONCH+UNI\_MLP & 0.82 & 0.87 & 0.78 & 0.78 \\
        \hline
        Hist2Cell      & 0.81 & 0.87 & 0.77 & 0.77 \\
        \hline
    \end{tabular}}%
    }
\end{table}
\subsection{Generalization to External Datasets}
\label{sec:generalization}

A key goal of our model is to generalize to new histology images without gene-spot expression. To test this, we train the MLP on her2st and evaluate on STNet, and vice versa.

\cref{tab:l1_CC_unseen} shows that our method with concatenated CONCH+UNI embeddings consistently outperforms others, achieving the lowest L1 and highest CC scores in both directions. In contrast, single-model embeddings sometimes underperform relative to Hist2Cell. These results highlight that combining foundation model embeddings improves generalization.
\begin{table}[t]
    \centering
    {
    \caption[Generalization across datasets]{\textbf{Evaluation of model generalization across datasets.} L1 error (lower is better) and CC score (higher is better) for cross-dataset predictions.}
    \label{tab:l1_CC_unseen}
    \centering
    \resizebox{
    0.8\columnwidth}{!}{%
    \renewcommand{\arraystretch}{1.3}
    \begin{tabular}{|l|c|c|c|c|}
        \hline
        \multirow{2}{*}{\textbf{Model}} & \multicolumn{2}{c|}{\textbf{her2st to STNet}} & \multicolumn{2}{c|}{\textbf{STNet to her2st}} \\
        \cline{2-5}
         & \textbf{L1 $\downarrow$} & \textbf{CC $\uparrow$} & \textbf{L1 $\downarrow$} & \textbf{CC $\uparrow$} \\
        \hline
        CONCH+UNI\_MLP & 0.57 & 0.32 & 0.55 & 0.40 \\
        \hline
        CONCH\_MLP     & 0.60 & 0.23 & 0.55 & 0.40 \\
        \hline
        UNI\_MLP       & 0.59 & 0.28 & 0.56 & 0.36 \\
        \hline
        Hist2Cell      & 0.77 & 0.27 & 0.61 & 0.40 \\
        \hline
    \end{tabular}}%
    }
\end{table}

\subsection{Computational Efficiency}
\label{sec:computational_complexity}

Our method delivers comparable or superior performance to Hist2Cell while substantially reducing computational overhead. \cref{tab:computation_complexity} reports GPU memory usage and training times for different embedding configurations (CONCH, UNI, and CONCH+UNI) relative to Hist2Cell.

During training, our approach reduces GPU memory consumption by approximately 43GB and shortens average training time by over 4.5 hours per run. These gains are consistent across leave-one-out experiments on both the STNet and her2st datasets, as well as full dataset training. This efficiency underscores the practicality of our method for large-scale deployment.

\begin{table}[t]

    \caption[Computational efficiency comparison]{\textbf{Computational efficiency of different models.} Comparison of GPU usage and training times across datasets. 1: Leave-one-out on STNet, 2: Leave-one-out on her2st, 3: Cross-dataset: Train on her2st, evaluate on STNet, 4: Cross-dataset: Train on STNet, evaluate on her2st.}
\label{tab:computation_complexity}
    \centering
    \resizebox{\columnwidth}{!}{%
    \renewcommand{\arraystretch}{1.8}
    \begin{tabular}{|l|c|c|c|c|c|}
        \hline
         & \textbf{GPU Usage} & \textbf{Time 1} & \textbf{Time 2} & \textbf{Time 3} & \textbf{Time 4} \\
        \hline
        CONCH+UNI\_MLP & 570MB  & 6m 5s & 9m 13s & 82m 15s & 58m 47s \\
        \hline
        CONCH\_MLP     & 546MB & 6m 30s & 9m 3s  & 82m 16s & 58m 8s \\
        \hline
        UNI\_MLP       & 570MB & 6m 28s & 9m 25s  & 83m 37s & 58m 13s \\
        \hline
        Hist2Cell      & 44951MB & 383m 24s & 171m 39s & 280m 1s  & 420m 34s \\
        \hline
    \end{tabular}}
    
\end{table}

\section{Ablation Study}

We conducted an ablation study on the her2st and STNet datasets to assess whether adding a third foundation model, UNI2, improves the performance on predicting cell-type proportions. UNI2 produces 1536-dimensional embeddings, equivalent in size to the concatenation of CONCH (512D) and UNI (1024D), and is trained on 200M pathology images using a ViT-h/14 architecture.

\begin{figure}[ht]
    \centering
    { 
    \includegraphics[width=\columnwidth]{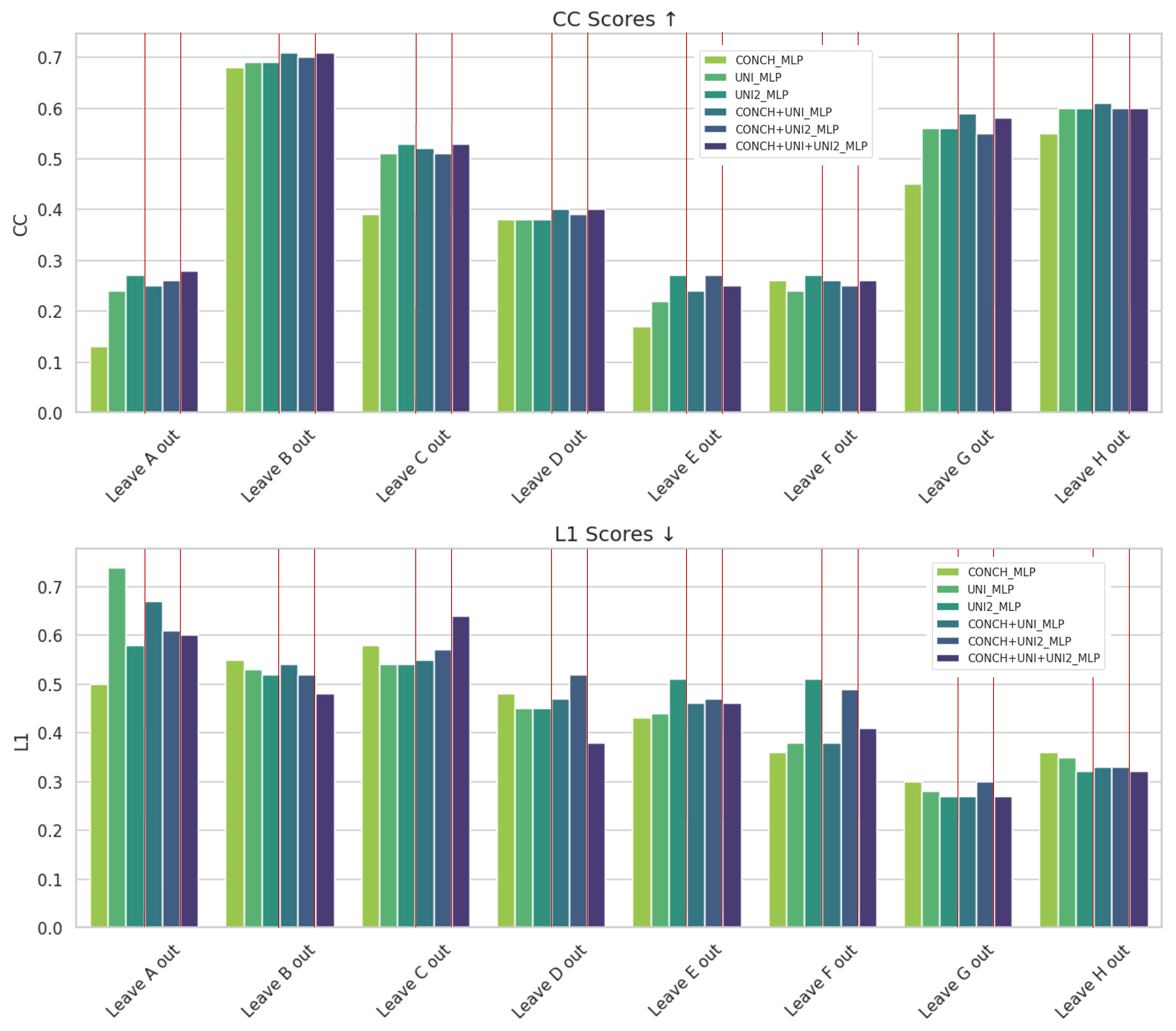}
    \caption[Ablation study on her2st dataset]{\textbf{Ablation study on the her2st dataset.} CC and L1 scores for different embedding configurations in a leave-one-out experiment on the her2st dataset.}
    \label{fig:albation_her2st}}
\end{figure}

\cref{fig:albation_her2st,fig:albation_stnet} show CC and L1 scores across configurations: single-model (CONCH, UNI, UNI2), two-model (CONCH+UNI, CONCH+UNI2), and three-model (CONCH+UNI+UNI2). In many cases, UNI2 alone or the CONCH+UNI combination performs best, indicating that careful selection of embedding sources is more important than simply increasing dimensionality.

\section{Discussion}
\label{sec:aim3_discussion_summary}
We present a lightweight approach for predicting cellular composition directly from H\&E-stained histology images, leveraging embeddings from pre-trained pathology foundation models and a simple MLP regressor. Our method consistently matches or surpasses Hist2Cell in both cell-type abundance prediction and colocalization analysis. While concatenating embeddings enhances performance, our ablation study shows that selecting optimal combinations is more effective than simply increasing dimensionality.

The model generalizes well to unseen datasets and significantly reduces GPU memory usage and training time compared to Hist2Cell, improving scalability and practicality. Overall, this study demonstrates the feasibility of using pre-trained computational pathology models to efficiently and accurately infer cellular composition without ST data, supporting broader application of pathology foundation models in clinical and translational research.

\section*{Impact Statement}
This paper presents work whose goal is to advance the field of Machine Learning. There are many potential societal consequences of our work, none which we feel must be specifically highlighted here.

\bibliography{example_paper}
\bibliographystyle{icml2025}

\clearpage
\onecolumn
\renewcommand{\thefigure}{A\arabic{figure}}
\setcounter{figure}{0}

\section*{Appendix}

\begin{samepage}
\begin{figure}[!htb]
    \centering
    {\scriptsize
    \includegraphics[width=0.8\textwidth]{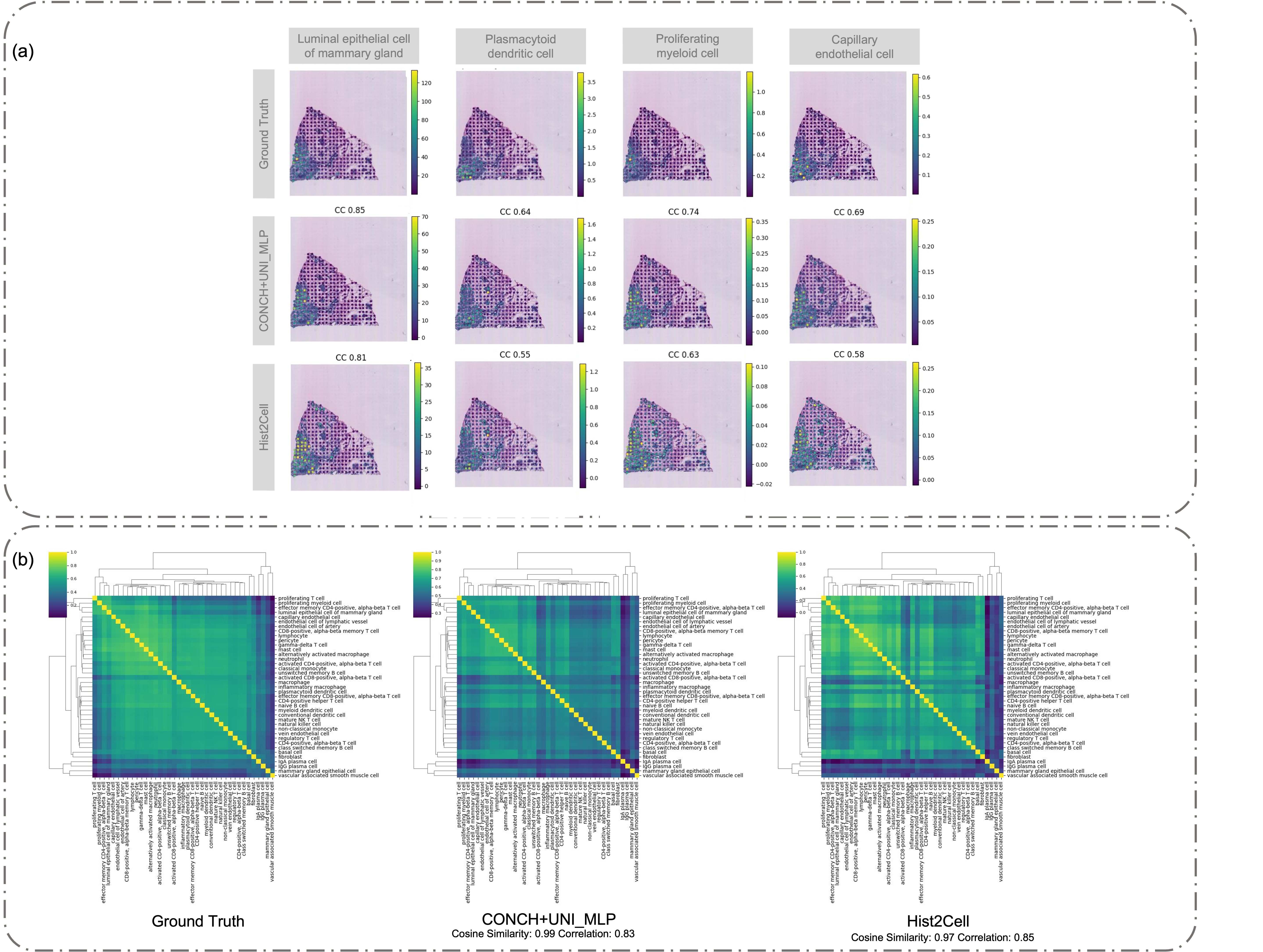}
    
    \caption[Visualization of cell-type predictions and colocalization]{\textbf{Visualization of cell-type predictions and colocalization.} (\textbf{a}) Predicted cell-type distributions for a representative slide, comparing ground truth, CONCH+UNI\_MLP, and Hist2Cell. (\textbf{b}) Clustermaps of bivariate Moran’s R for spatial colocalization.}
    \label{fig:B1}
    }
    \vspace{-10mm}
\end{figure}
\vspace{-10mm}
\begin{figure}[!tb]
\vspace{-10mm}
    \centering
    {\scriptsize 
    \includegraphics[width=0.6\columnwidth]{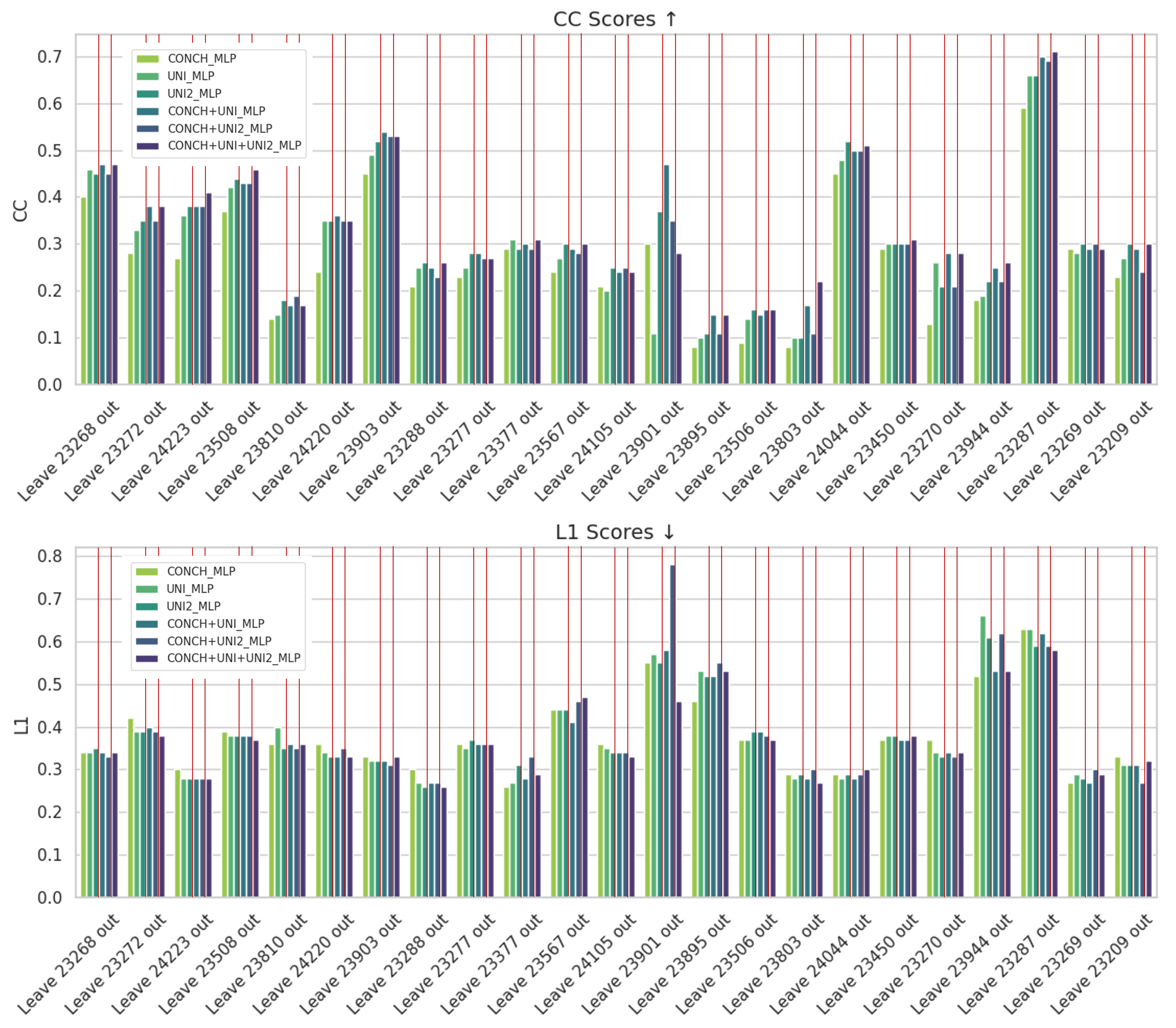}
    \caption[Ablation study on STNet dataset]{\textbf{Ablation study on the STNet dataset.} CC and L1 scores for different embedding configurations in a leave-one-out experiment on the STNet dataset.}
    \label{fig:albation_stnet}}
\end{figure}
\end{samepage}
\end{document}